\documentclass[conference]{IEEEtran}
\IEEEoverridecommandlockouts
\usepackage{cite}
\usepackage{amsmath,amssymb,amsfonts}
\usepackage{algorithmic}
\usepackage{graphicx}
\usepackage{textcomp}
\usepackage{csquotes}
\usepackage{xcolor}
\def\BibTeX{{\rm B\kern-.05em{\sc i\kern-.025em b}\kern-.08em
    T\kern-.1667em\lower.7ex\hbox{E}\kern-.125emX}}

\usepackage{fancyhdr}
\begin{document}

\title{What's Next in Affective Modeling? Large Language Models}

\author{\IEEEauthorblockN{Nutchanon Yongsatianchot}
\IEEEauthorblockA{\textit{Khoury College of Computer Science} \\
\textit{Northeastern University}\\
Massachusetts, USA \\
nutjung.nutlc@gmail.com}
\and
\IEEEauthorblockN{Tobias Thejll-Madsen}
\IEEEauthorblockA{\textit{The School of Psychology \& Neuroscience} \\
\textit{University of Glasgow}\\
Glasgow, Scotland \\
t.thejll-madsen.1@research.gla.ac.uk }
\and
\IEEEauthorblockN{Stacy Marsella}
\IEEEauthorblockA{\textit{Khoury College of Computer Science} \\
\textit{Northeastern University}\\
Massachusetts, USA \\
s.marsella@northeastern.edu}
}

\maketitle
\thispagestyle{fancy}

\begin{abstract}
Large Language Models (LLM) have recently been shown to perform well at various tasks from language understanding, reasoning, storytelling, and information search to theory of mind. In an extension of this work, we explore the ability of GPT-4 to solve tasks related to emotion prediction. GPT-4 performs well across multiple emotion tasks; it can distinguish emotion theories and come up with emotional stories. We show that by prompting GPT-4 to identify key factors of an emotional experience, it is able to manipulate the emotional intensity of its own stories. 
Furthermore, we explore GPT-4's ability on reverse appraisals by asking it to predict either the goal, belief, or emotion of a person using the other two. In general, GPT-4 can make the correct inferences. We suggest that LLMs could play an important role in affective modeling; however, they will not fully replace works that attempt to model the mechanisms underlying emotion-related processes. 
\end{abstract}


\section{Introduction}

Large language models (LLM), such as ChatGPT or GPT-4 from OpenAI, have demonstrated an impressive range of abilities from answering scientific, medical, and legal questions to reasoning, programming, writing poems, songs, and stories \cite{brown2020language, openai2023technical, bubeck2023sparks}. With the ability to process language, interact with humans, and generate various stories, it is natural to ask about their emotion prediction abilities and whether or not they could serve as or be a part of an affective model - predicting and explaining emotions and reactions to emotions. 

Research has already begun to study and demonstrate different abilities of LLMs associated with human psychology and cognition, such as decision-making, causal reasoning, theory of mind, and emotion understanding \cite{binz2023using,bubeck2023sparks,kosinski2023theory}. For instance, Binz and Shultz \cite{binz2023using} used common psychological tests to study the decision-making, information search, and causal reasoning skills of GPT-3. They found that it can solve most of these tasks to a human level or better, and GPT-3 also displays some common human biases. Kosinksi (2023) evaluated the ability of OpenAI's GPTs to solve Theory of Mind (ToM) tasks and showed that recent models such as ChatGPT and GPT-4 could solve most of these problems \cite{kosinski2023theory}. Furthermore, Bubeck et al. (2023) found that GPT-4 outperforms ChatGPT on ToM and emotion understanding \cite{bubeck2023sparks}. While testing the models on various topics such as coding, mathematics, and psychology, they only tested the models on a few examples and did not evaluate the psychological aspects of emotion in detail. This current work seeks to further test and demonstrate an LLM's ability on a range of emotion-related tasks involving predicting and explaining emotion-related phenomena. 

This work focuses only on GPT-4, an LLM by OpenAI publicly available on March 2023. This is because OpenAI provides an easy access interface for testing as well as an API and, more importantly, at the time of writing, GPT-4 is the most powerful model \cite{peng2023instruction, LMSYS2023chatbot}. GPT-4 is an autoregressive decoder-only transformer-based large language model \cite{brown2020language, openai2023technical}. LLMs are pretrained on a large corpus of datasets to predict the next word and fine-tuned using Reinforcement Learning with Human Feedback to make the models able to follow instructions while being safe and aligning with human values \cite{ouyang2022training}.

Therefore, in this work, we aim to further investigate GPT-4's ability to predict human emotion and related elements, including beliefs and goals, from the perspective of computational modeling of emotions. In particular, we subject it to various tests, some of which are based on tasks proposed in the field to evaluate emotion models. First, we start by testing GPT-4's knowledge of emotion, including emotion theories, generating emotional stories, identifying important factors influencing emotion, and creating and annotating emotion stories using those factors. 
Then, we explore its ability on reverse appraisal, specifically given two of a person's goals, beliefs, or emotions we see if it can predict the remaining one \cite{hareli2010emotional, de2014reading, ong2019computational}.

We found that GPT-4 can do all these tasks remarkably well. In particular, it can generate diverse emotional stories, identify and annotate key factors, explain emotional events, and do reverse appraisal when given enough context. Our results serve as preliminary tests demonstrating GPT-4's ability to predict emotion and related features and serve as an affective model. 

\section{Dialog with GPT-4}

To begin, we look at how GPT-4 answers questions related to emotion theory and how it can use this information to generate various emotion stories.
First, we show that when prompted to explain contrasting emotion theories, GPT-4 is able to report key defining characteristics \footnote{The knowledge cutoff for GPT-4 is September 2021, and therefore any development in emotion theories since that date is not available to the model.}.  While this does not challenge the ability to make emotional inferences, it highlights that the model can correctly produce declarative knowledge about emotions and emotion theories. We ask GPT-4 as follows.
\begin{itemize}
    \item Explain Ekman’s emotion theory in 3 - 4 sentences.
    \item \textbf{GPT-4}: Ekman's emotion theory, also known as the "basic emotion theory," posits that there are six universal emotions experienced by all humans, regardless of culture or background. These emotions are happiness, sadness, fear, anger, disgust, and surprise. The theory also suggests that these emotions are innate, hardwired in the brain, and can be recognized through distinct facial expressions and physiological responses, allowing for efficient communication of emotional states between individuals.
    \item Explain Barrett’s emotion theory in 3 - 4 sentences.
    \item \textbf{GPT-4}: Barrett's emotion theory, also known as the theory of constructed emotion, posits that emotions are not universal, hardwired biological responses, but rather constructed experiences based on an individual's past experiences, culture, and context. Developed by psychologist Lisa Feldman Barrett, this theory suggests that the brain creates emotions by categorizing and making sense of the physical sensations and information it receives from the body and the environment. In essence, emotions are not innate, but instead are flexible, ever-changing constructs that vary from person to person and from one situation to another.
    \item Explain Appraisal theory of emotion in 3 - 4 sentences.
    \item \textbf{GPT-4}: Appraisal theory of emotion suggests that emotions are determined by our individual evaluations or interpretations of a particular event or situation. It posits that our emotional experiences arise from cognitive processes, specifically the way we appraise and assess events based on their personal relevance, goal congruence, and coping potential. Essentially, appraisal theory emphasizes the role of cognitive processes in shaping our emotional experiences, as different individuals may experience different emotions in response to the same situation due to differing appraisals.
\end{itemize}
Each of GPT-4's responses captures well a high-level description of basic emotion theory \cite{ekman1999basic}, theory of constructed emotions \cite{barrett2017theory}, and appraisal theory \cite{lazarus1991emotion}. Given that the responses are prompted to be quite short they lack nuance, but show that GPT-4 can describe well three different emotion theories. It was expected that GPT-4 would do well summarising theories as it does not require any capability to reason emotionally.

To further test its ability on working with emotions, we ask GPT-4 to generate short stories for five different emotions (joy, anger, sadness, fear, and guilt). Storytelling is a great tool to explore emotional ability as they allow expression of emotions \cite{mckee_storytelling_2003}, and use emotions to connect characters, their motivations, goals, and achievements \cite{vonnegut_palm_1981}. Oatley (1999) argues the importance of fictional stories in emotion research. He makes the comparison between fictional stories and mental simulation; in essence, to create and understand stories, both the writer and the reader need knowledge about the world, its agents, their goals, and motivations \cite{oatley_why_1999}. Therefore, by using stories as test cases, we get a better sense of GPT-4's ability to create the stories' characters, human social interaction, social and emotional inference, and theory of mind \cite{mar_function_2008}. To get GPT-4 to create stories, we prompt it as follows.
\begin{itemize}
    \item Generate 5 emotional short 2 - 3 sentences scenarios for joy, anger, sadness, fear, and guilt.
    \item \textbf{GPT-4}: 1. Joy: As she opened the envelope, her eyes lit up and her heart filled with joy. She had finally received acceptance into her dream university, making all her hard work pay off. 
    2. Anger: He slammed his fist on the table, his face red with anger. His roommate had used up the last of his favorite coffee without replacing it, ruining his morning ritual.
    3. Sadness: She sat at the edge of her bed, tears streaming down her face. As she clutched her grandmother's shawl, the realization that she was truly gone began to sink in.
    4. Fear: His heart raced as he walked down the dark alley, the sound of footsteps growing louder behind him. Fear gripped him tightly, leaving him paralyzed and unable to run.
    5. Guilt: She stared at the shattered vase on the floor, her mother's prized possession now in pieces. Guilt washed over her as she realized she could never replace what she had carelessly destroyed.
\end{itemize}
As can be seen from these examples, GPT-4 can generate short stories that capture the key relationship between a person, their environment, and an emotion. For instance, GPT-4's stories show that joy can arise from meeting positive valenced goals, sadness can come from loss, and fear from anticipating potential harm, all of which align with emotional theories. 

However, to better explore how well GPT-4 works with emotions, we wanted to see its ability to use factors that influence emotional experiences. Therefore, we asked GPT-4 to provide us with five factors that could influence emotional experiences and their intensity. After the fist five, we asked for an additional five. The model output has been shortened to fit this paper. Full prompts and answers are available on GitHub (github.com/yongsa-nut/ModelingWSACII23). 
\begin{itemize}
    \item List the top 5 factors that can influence emotion intensity. 
    \item \textbf{GPT-4}: 1. Personality: An individual's personality plays a significant role in determining the intensity of emotions they experience... 
    2. Situation or context: The specific situation or context in which an emotion is experienced can greatly influence its intensity... 
    3. Cognitive appraisal: The way an individual interprets or evaluates an event or situation can also influence the intensity of their emotions...
    4. Physiological arousal: The level of physiological arousal or activation in the body can impact the intensity of emotions experienced...
    5. Social and cultural factors: Social and cultural factors can also play a significant role in shaping the intensity of emotions experienced by individuals...
    \item List another 5.
    \item \textbf{GPT-4}: 6. Emotional history and past experiences...
    7. Mood and emotional state...
    8. Age and developmental factors...
    9. Attachment style...
    10. Expectations and beliefs...
\end{itemize}

We now wanted to test GPT-4's ability to use this information in story generation. To display a meaningful application of emotional theory, GPT-4 must then be able to use the factors highlighted to moderate the intensity of emotional stories.  To test this, we prompted GPT-4 as follows.
\begin{itemize}
    \item Now generate 3 joyful short scenarios ranging from low to high intensity and add the factors in parentheses when they are first mentioned.
    \item \textbf{GPT-4}: Low intensity joyful scenario: You're walking through a park on a sunny day, feeling content and relaxed (mood and emotional state). As you pass by a beautiful flower bed, you notice a butterfly landing on a vibrant flower. The sight of this small, delicate creature brings a gentle smile to your face, as you remember your childhood fascination with butterflies (emotional history and past experiences). You take a moment to appreciate the simple beauty of nature before continuing your leisurely walk. 
    \item \textbf{GPT-4}: Medium intensity joyful scenario:
    You're at a friend's birthday party, surrounded by your close friends and family (social and cultural factors). As you all sing "Happy Birthday" to your friend, you begin to feel a warm sense of happiness and connection with the people around you (attachment style). Your friend's face lights up with joy as they blow out the candles on their cake, and you can't help but feel excited and happy for them (cognitive appraisal). The overall atmosphere of the party lifts your spirits, and you find yourself genuinely enjoying the celebration.
    \item \textbf{GPT-4}: High intensity joyful scenario: After months of hard work and dedication, you finally receive the news that you've been accepted into your dream university (situation or context). Your heart races with excitement (physiological arousal), and you can hardly believe that your dreams are coming true (expectations and beliefs). You feel an overwhelming sense of joy and pride, knowing that your efforts have paid off (cognitive appraisal). As you share the news with your family and friends, their reactions of happiness and congratulations only amplify your own feelings of elation (social and cultural factors). This moment of achievement marks a significant milestone in your life, and the joy you feel is incredibly intense and memorable (age and developmental factors).

\end{itemize}
As can be seen from these scenarios, the model is able to change the emotional intensity and highlight when different factors that influence emotional experiences are present in the scenarios. To further challenge the model, we asked it to turn the low-intensity scenario into a high-intensity one.  

\begin{itemize}
    \item Can you turn the low-intensity scenario into a high-intensity one?
    \item \textbf{GPT-4}: You're walking through a park on a sunny day, feeling content and relaxed (mood and emotional state). As you pass by a beautiful flower bed, you notice a rare species of butterfly landing on a vibrant flower – the very species you've been searching for as part of your research project (situation or context). Your heart races with excitement (physiological arousal) as you realize the significance of this discovery (cognitive appraisal). This moment not only sparks your childhood fascination with butterflies (emotional history and past experiences) but also carries great importance for your academic career and passion for environmental conservation (expectations and beliefs). You quickly take photos and document the butterfly's location, knowing that this find will greatly contribute to your research and potentially lead to new conservation initiatives (cognitive appraisal). As you share the news with your colleagues and mentors, their reactions of amazement and congratulations (social and cultural factors) only amplify your own feelings of elation and accomplishment. This incredible encounter transforms your leisurely walk in the park into an intensely joyful and memorable experience.
\end{itemize}

One key observation is that to make a scenario more emotionally intense, the model increases both the number of factors in the scenario and the general importance of the situational context (e.g., causal walk in a park vs. looking for butterflies for a research project). The story is still quite simple, but by using simple building blocks such as the fact that emotions are both a social and personal experience, GPT-4 manages to manipulate the perceived emotional intensity of the story. We also prompted GPT-4 to generate scenarios with different intensities for fear with similar results (See GitHub). 

Lastly, we prompted it to generate a list of five factors from Apprisal Theory and then generate an emotional story with the factors annotated. 

\begin{itemize}
    \item List the five main factors in appraisal theory of emotions
    \item \textbf{GPT-4}:  1. Relevance: This refers to how relevant the event or situation is to the individual's personal goals or well-being. 2. Implications: This assesses the potential consequences of the event or situation. 3. Coping Potential: This evaluates the individual's ability to cope with or manage the event or situation. 4. Normative Significance: This involves evaluation of the event or situation based on social norms or values. 5. Future Expectancy: This assesses the anticipated outcome of the event or situation. 
    \item Write a 4 -5 sentences joyful scenario and add the appraisal factors above in parentheses when they are mentioned
    \item \textbf{GPT-4}: Sophia just found out that she passed her final exams with flying colors, exceeding her own expectations (Implications). She feels a rush of joy knowing that her hard work paid off and that she's one step closer to her dream of becoming a lawyer (Relevance). She also feels confident that she's capable of handling the challenges of law school (Coping Potential). Her parents are incredibly proud, reinforcing her belief that she's on the right path (Normative Significance). Now, she looks forward to the next phase of her education journey with excitement and anticipation (Future Expectancy).
\end{itemize}

We observe that the factors provided by GPT-4 are based on Scherer's model of appraisal theory, the Component Process Model, and GPT-4 can appropriately annotate the scenario using these factors\cite{scherer2001appraisal}. This shows a potential use case of GPT-4 to both annotate and potentially explain existing stories and to create new ones grounded in a specific emotion theory. Next, we turn to the question of whether or not it can predict the missing appraisal factors given other factors, the problem of reversed appraisal.

\section{Reverse Appraisal}

In this section, we follow the line of work on Reverse Engineering Appraisal or Reverse Appraisal to test GPT-4 \cite{hareli2010emotional, de2014reading, hareli2019reverse}. From an appraisal theory perspective, emotion arises from subjective evaluation based on the person's goals and beliefs \cite{arnold1960emotion, lazarus1991emotion, moors2013appraisal}. Therefore, given two of the three factors (goal, belief, and emotion), we should be able to reverse engineer the appraisal to predict the remaining one. For instance, if we are in a sporting event between two teams and we see our friends happy with the result, we can predict that they are rooting for the winning team. 

There has been a long line of work using the Bayesian framework to do reverse appraisal \cite{de2014reading, wu2014joint, yongsatianchot2016integrating, houlihan2022reasoning} (see \cite{ong2019computational} for a review). One key limitation of the Bayesian framework, common in other existing approaches too, is that one needs to map the description of the event into a suitable format, such as prior and likelihood distributions. This also limits the generalization of the approach to apply to any arbitrary domain. Here, we demonstrate the power of GPT-4 for this task.

In this work, our scenarios are based on team sports as they provide a clear structure for goals (which team to support) and beliefs (the outcome of the game). We consider two sporting competitions, one real (football/soccer) and one fictional (Quidditch). Quidditch is a fictional team sport in the Harry Potter series \cite{wiki2023Quidditch} (it has been adapted into a real-life version with different a ruleset to suit real-world constraints). Here, we will use the fictional rules for the prompts. Each team consists of 4 roles: Chasers (scoring goals worth 10 points), Beater (defending the team), Keeper (blocking shots from Chasers), and Seeker (catching the golden snitch). The game ends when the Seeker catches the golden snitch yielding 150 points to the Seeker's team. Even though the golden snitch yields a significant amount of points, it is possible that the team that captures it can still lose.

\subsection{Predicting Goals}

First, we ask GPT-4 to predict the goal of a person in the audience, given beliefs and emotions. The prompt is as follows: 

\begin{displayquote}
Consider the following scenario. Teams A, B, and C are competing in a football league with 17 other teams. Today's match is Team A vs. Team B. Robert is attending the match today. The match begins. 

\emph{Team A is attacking and getting in a good position to score a goal.}
\end{displayquote}

The questions include: 
\begin{itemize}
    \item If Robert was \{hopeful, anxious, disappointed, relieved, happy, sad, neutral\}, what team does he support? 
    \item If Robert was \{mad, annoyed\}, what team does he support?
    \item If Robert was angry at \{the attacker who missed the shot, the goalie, Chaser, Keeper\}, what team does he support?
    \item If Robert was proud of \{the goalie, the shooter, Chaser, Keeper\}, what team does he support?
\end{itemize}

A full list of questions as well as responses from GPT-4 can be found on GitHub. Effectively, the questions can be grouped into two categories (following the OCC theory of emotions roughly): emotions focused mainly on the event and emotions focused on the agent \cite{ortony1988cognitive}. When we prompt, we ask all the questions in the same group at the same time. Along the same line, we consider three types of events (emphasized in the prompt above): 1) attacking about to score a goal, 2) attempting to score a goal but failed, and 3) successfully scoring a goal. 

\begin{figure}[htbp]
\centering
\includegraphics[scale=0.5]{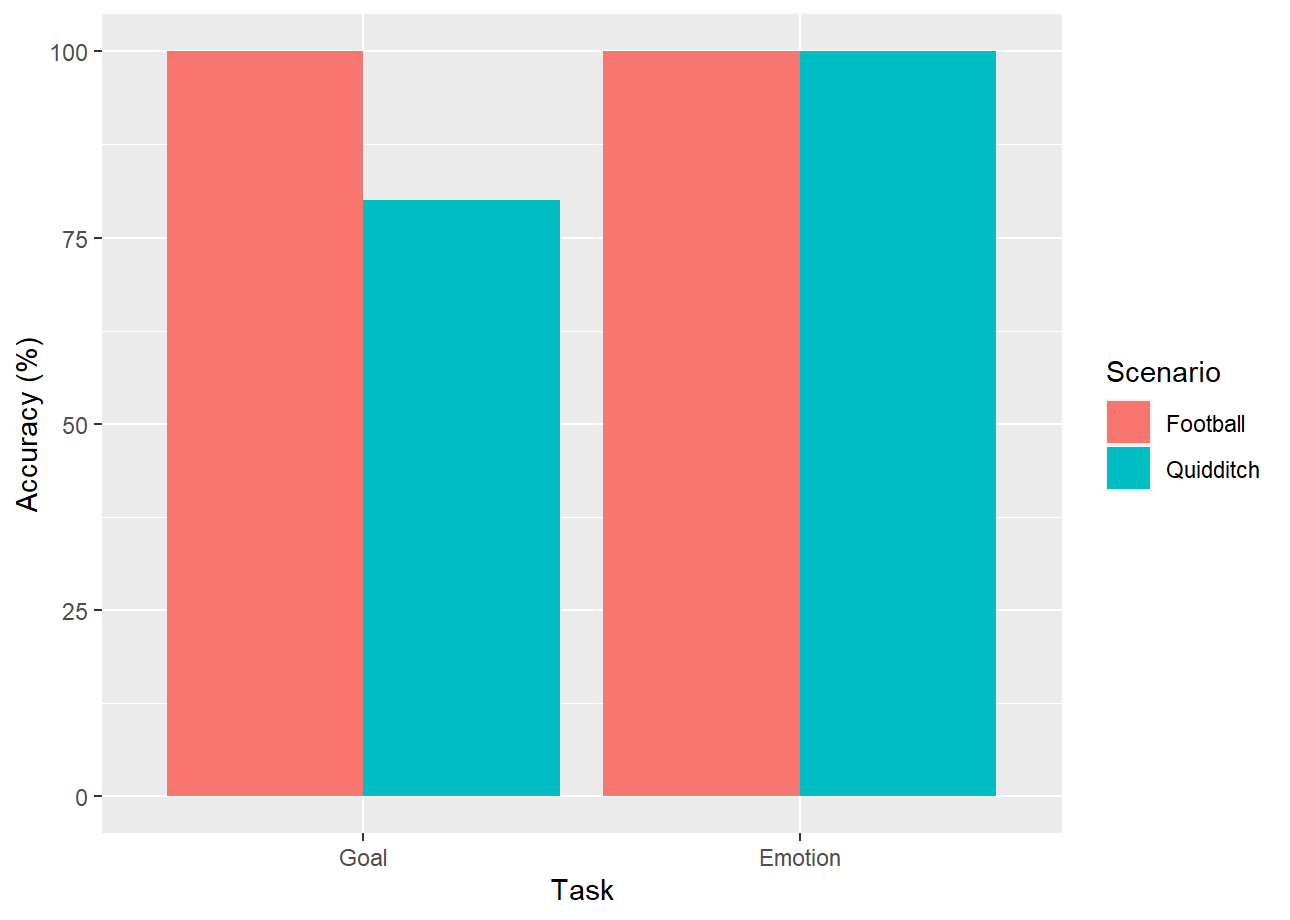}
\caption{Prediction Results for Goal and Emotion}
\label{fig:results}
\end{figure}

Figure \ref{fig:results} shows the results. For the football scenario, there are 22 questions in total and GPT-4 got all of them correctly. 
For the quidditch scenario, we expanded the third event type into two more events: catching the golden snitch while behind by 50 points (results in winning) and while behind by 160 points (results in losing). In total, there are 30 questions. 
GPT-4 got eight questions wrong (accuracy = 80\%). In particular, in the event that the chaser is about to score a goal, GPT-4 answers that Robert could support either team for four questions. Additionally, in the event that the seeker caught the golden snitch but was behind by 160 points resulting in losing, GPT-4 got another four questions (disappointed, relieved, happy, sad) wrong, and for the other questions for this event, GPT-4 got them right but offered a wrong explanation. For example, "If Robert was proud of the Seeker, he likely supports Team A, as their Seeker successfully caught the golden snitch, giving them a chance to win the match." This demonstrates that GPT-4 does not get the context of Quidditch quite right. Therefore, we follow up by asking it about the points that the golden snitch gives and GPT-4 can answer it correctly (150). Then, we asked it to use this information to revise the previous answers. This time GPT-4 answered all the questions correctly. This shows that once GPT-4 has enough information on the context, it could use it to make the right predictions.   

As alluded to earlier, GPT-4 can also offer an explanation of why Robert may feel a certain emotion. Here are a few examples. "If Robert was angry at the goalie who couldn’t block the goal, he likely supports Team B, as their goalie failed to save the goal," and "If Robert was angry at the defense as he couldn’t intercept the play earlier, he likely supports Team B, as their defense allowed Team A to get a chance to score." 

Lastly, we observed that GPT-4's answers can depend on the context, specifically the questions that are asked at the same time. Consider the following question, "If Robert were mad, what team does he support?" When we ask this question together with other agent-related questions, GPT-4 answered correctly. However, when we ask this question alone, GPT-4 responds that it cannot determine which team Robert supports based on the information provided. 

\subsection{Predicting Emotions}

Next, we provide GPT-4 with the same prompts, but this time we tell it Robert's goal ("He is a big fan of Team A") and the outcome of the event and ask for emotions. Here are some examples of the events. 

\begin{itemize}
    \item Team \{A, B\} \{missed a shot, scored\}.
    \item Team \{A, B\} scored an owned goal. 
    \item Team \{A, B\} scored a bicycle kick goal. 
    \item Team \{A, B\}’s seeker catches a golden snitch while Team A’s behind by \{100, 160\} points. 
\end{itemize}

Figure \ref{fig:results} show the results on predicting emotions. In short, for both football (14 questions) and Quidditch (10 questions) scenarios, GPT-4 answered all the questions correctly based on our manual evaluation in terms of possibility. Below are a few selected examples: 

\begin{itemize}
    \item Robert would feel happy or excited that Team B scored an own goal, as it benefits Team A.
    \item Robert would likely feel impressed but disappointed or concerned if Team B scored a bicycle kick goal, as it's an amazing play but hurts his favorite team's chances of winning.
    \item If Team A's seeker catches the golden snitch while they're behind by 100 points, Robert would feel excited and hopeful, as catching the snitch awards 150 points and would give Team A the win.
    \item If Team A's seeker catches the golden snitch while they're behind by 160 points, Robert would feel disappointed, as even though they caught the snitch, they would still lose the match. 
\end{itemize}

\subsection{Predicting Beliefs}

Lastly, we provide GPT-4 with the same prompts, except this time we tell it Robert's goal ("He is a big fan of Team A") and his emotion and now ask it to generate three possible outcomes as follows: "If Robert feels relieved/disappointed/happy/sad/extremely pleased/angry, list three possible events that could have happened in the game." 

In all cases, GPT-4 can generate reasonable and appropriate events. We present here some of the outputs and the complete responses can be found on GitHub. The first three are the football scenario, and the last three are the Quidditch scenario.

\begin{itemize}
    \item If Robert feels relieved, three possible events that could have happened in the game are: 1. Team A scored a last-minute goal to tie or win the match. 2. Team A's goalkeeper made a crucial save to prevent Team B from scoring. 3. Team A successfully defended against a dangerous attack from Team B.
    \item If Robert feels happy, three possible events that could have happened in the game are: 1. Team A won the match against Team B. 2. Team A played exceptionally well throughout the match. 3. Team A scored a beautiful or exciting goal.
    \item  If Robert feels extremely pleased, three possible events that could have happened in the game are: 1. Team A won the match by a large margin or scored a significant number of goals. 2. Team A secured a spot in the playoffs or won a championship. 3. A player from Team A achieved a personal milestone, like breaking a scoring record or receiving an award.
    \item If Robert feels relieved, three possible events that could have happened in the game are: 1. Team A made a last-minute score to win the game. 2. Team A's goalkeeper successfully blocked a crucial shot from Team B. 3. Team A caught the Snitch just before Team B could, securing their victory.
    \item  If Robert feels happy, three possible events that could have happened in the game are: 1. Team A won the match against Team B. 2. Team A's players performed exceptionally well, showcasing their skills. 3. Team A caught the Snitch, securing a victory.
    \item If Robert feels extremely pleased, three possible events that could have happened in the game are: 1. Team A won the match by a large margin, showcasing their dominance. 2. Team A's star player scored a record-breaking number of points. 3. Team A's victory propelled them to the top of the league standings.
\end{itemize}

Interestingly, we also observe that GPT-4 can still be sensitive to the context, similar to the case of predicting goals. In particular, when we ask only GPT-4 to consider when Roberts feels extremely pleased, here are the three possible events: 
\begin{itemize}
    \item Team A scores a goal, taking the lead in the game.
    \item Team A’s goalkeeper makes an incredible save, preventing Team B from scoring.
    \item Team A’s star player performs a stunning play or skill move, existing the crowd and Robert.
\end{itemize}

We see that these are different from the earlier responses, where the intensity of positive emotions is likely to be lower in these responses than in the earlier responses.

\section{Discussion}

In this work, we present multiple emotion-related tasks that an LLM, specifically GPT-4, can do, from explaining emotion theories, listing important factors, generating stories, explaining evolving stories, and reverse appraisal predicting goals, beliefs, or emotions given the other two. We see that GPT-4 is capable of doing all these tasks exceptionally well. 

Still, what we have shown here is simply preliminary tests on emotion perception and prediction. Much work needs to be done to fully understand the scope and limits of LLMs. Nevertheless, we believe that these early results show that GPT-4 far exceeds any existing computational models in terms of predictions across a wide range of tasks, the ability to explain the results, and the ease of use and readiness to apply to any domain and application. The last point is perhaps the most crucial ability that LLMs have over most, if not all, existing computational affective models, where modelers need to fit, enhance, and/or specify the model to the context. In LLMs, the context is the basis of the model, and they are also equipped with vast amounts of knowledge - not only is it an enormous advantage in terms of building applications, but it is also consistent with views of emotion that argue for such contextualization, such as appraisal theory and theory of constructed emotion \cite{barrett2017theory, moors2013appraisal}. Still, we observe that GPT-4 can be sensitive to prompts and careful prompting is still needed to make it work well. Additionally, one crucial weakness of GPT-4, in contrast to existing models, is that we do not fully understand how it arrives at the answers besides learning statistical regulation in data to predict the next words.

In this work, our evaluation can be viewed as involving story generation and roleplaying characters. Story generation, roleplaying, and perspective-taking are great test cases for the capabilities to predict and generate emotionally appropriated outputs because all of these tasks require a simulation of social processes and internal states of the agents \cite{mar_function_2008, oatley_fiction_2016}. As shown by authors before us \cite{kosinski2023theory}, LLMs are doing well at answering theory of mind tasks, and this paper adds additional evidence that GPT-4 can both generate and explain emotional scenarios we associate with theory of mind. 

On the flip side, the view on story generation also suggests a potential explanation on why LLMs can achieve these capabilities. These models have been trained on an enormous amount of literature and stories. In addition, their data also include internet data involving numerous human interactions; each can be considered a story in and of itself. Being able to predict these types of data well may have made them acquire the ability to explain and predict emotions.

Taking this argument further and more speculatively, it is conceivable that they may have learned a representation of emotional events, similar to appraisal theory and appraisal dimensions. More specifically, in order to predict next words in emotional contexts or stories well, the model may have learned a useful representation for this task. In this work, we have shown that they can do reverse appraisal when the context is sufficiently given without having to explicitly state appraisal theory. However, appraisal theory is still just a theory based on limited empirical data compared to LLMs. This suggests that LLMs could have learned richer, different appraisal dimensions for representing social and emotional events. Future work is needed to further explore the underlying representation that LLMs may have used to predict emotional events, which could lead to improving our appraisal theory, other emotional theories, and our understanding of emotion in general.

Furthermore, when specifically prompting GPT-4 to identify factors that impact the emotional intensity of stories, GPT-4 was able to show where in its stories it was using different factors and effectively used factors to change the emotional intensity of its generated stories. This result suggests that prompting GPT-4 to identify factors can be a useful way of doing prompt engineering to utilize the LLMs ability to describe and create emotional scenarios. That is, explicitly prompting GPT-4 to identify these factors provided an easy way of instructing it to change the emotional intensity of the story. This work only prompted factors relating to intensity provided by GPT-4 itself, but further work could consider exploring other factors such as relevance, duration of response, and so forth. Moreover, as tentatively shown, researchers can instruct GPT-4 to consider factors relevant to individual emotion theories and thereby either generate or analyze scenarios with a specific theory in mind. 

Our results, as well as many others, suggest that these LLMs are undeniably powerful and can solve various tasks, including emotion-related ones \cite{openai2023technical, bubeck2023sparks}. We also ask ourselves the obvious question: what is next? We consider here two possible objectives of affective models: 1) to advance our understanding of human emotion and the field of affective science and 2) to be part of virtual characters. For the first objective, we believe that LLMs will not replace the works that attempt to model the mechanisms underlying emotions and related processes, such as physiological, neurological, and cognitive processes. The model, like theory, is still needed to offer a deeper understanding of human emotions. Nevertheless, we believe LLMs will become a valuable tool for affective scientists and modelers as we now have a module that can process natural language in order to incorporate contexts effectively into the model. Furthermore, if the objective is to be part of virtual characters, we believe that LLMs will become the crucial component of affective processing that people will build upon, as some are already doing.

\section*{Ethical Impact Statement}

We seek to evaluate LLMs on their emotion understanding ability in this work. Several ethical problems are associated with LLMs. These include bias, harmful content, misinformation, privacy concerns, misuse by bad actors, and societal impact on employment. However, given how LLMs are going to be prevalent and impact us both researchers and the general population, it is crucial for research to explore and evaluate them to improve our understanding of them and their limits. 

\section*{Acknowledgment}
Tobias Thejll-Madsen is supported by the TTM: UK Research \& Innovation [EP/S02266X/1].

\bibliographystyle{ieeetr}
\bibliography{references}

\end{document}